\documentclass[10pt,twocolumn,letterpaper]{article}

\usepackage[pagenumbers]{cvpr} 

\definecolor{cvprblue}{rgb}{0.21,0.49,0.74}
\usepackage[pagebackref,breaklinks,colorlinks,allcolors=cvprblue]{hyperref}

\def\paperID{10967} 
\def\confName{CVPR}
\def\confYear{2025}

\title{Heterogeneous and Adept Snapshot Distillation for 3D Semantic Segmentation} 
\def\algorithmname{HAS-KD~}

\author{
  Xiaopei Wu\textsuperscript{1,2}, Yuenan Hou\textsuperscript{1 *}, Junkai Xu\textsuperscript{2}, Wenxiao Wang\textsuperscript{2}, Binbin Lin\textsuperscript{2 *}, Yu Li\textsuperscript{1}, Ping Li\textsuperscript{3} \\
  Haifeng Liu\textsuperscript{2}, Deng Cai\textsuperscript{2}, Wanli Ouyang\textsuperscript{1}
  \vspace{0.3em} \\
  \textsuperscript{1}Shanghai AI Laboratory \textsuperscript{2}Zhejiang University 
  \textsuperscript{3} Hangzhou Dianzi University \\
  * The corresponding authors
}

\begin{document}
\maketitle
\def\algorithmname{HAS-KD}


\begin{abstract}
Multi-modal fusion and multi-model ensembling are prevalent in enhancing the performance of 3D semantic segmentation. Despite the impressive performance, these methods either rely on auxiliary input signals or suffer from costly computational expense.
To efficaciously enhance the segmentation performance without introducing intolerable costs, we propose to transfer the rich knowledge from the multi-modal model (\ie, point clouds and images) and multiple model experts to the point-cloud-based network through knowledge distillation.
Specifically, we present Information-oriented Heterogeneous Distillation (IHD) to help the uni-modal model absorb the complementary knowledge from the multi-modal teacher. 
We design the Information-Oriented Filtering (IOF) strategy to select informative images 
from the continuous image sequence for multi-modal fusion. 
This practice can boost the performance of the multi-modal teacher, thus benefiting the learning of the student.
Besides, as opposed to vanilla model ensembling that requires the separate training of each expert, we propose Adept Snapshot Distillation (ASD). ASD treats the freely available model snapshots generated during the training phase as multiple experts, which significantly reduces the training cost for model ensembling. For each expert teacher, it only provides supervision to the student in the class where it is adept. 
The resulting Heterogeneous and Adept Snapshot Knowledge Distillation, dubbed \algorithmname, attains state-of-the-art results on ScanNetV2 and S3DIS datasets. \algorithmname~can be seamlessly integrated into contemporary 3D segmentation algorithms and bring considerable gains without introducing extra inference burdens. 
The code will be made publicly available upon publication.

\end{abstract}

\section{Introduction}
\label{sec:introduction}
3D semantic segmentation is a fundamental task in computer vision and plays an indispensable 
role in fine-grained 3D semantic scene understanding. Tremendous efforts have been made to achieve 
better segmentation performance by designing more powerful backbones~\cite{sparseconvnet,minkunet,ptv2,ptv3,swin3d,octformer}.

To further enhance the segmentation performance, many works resort to multi-modal fusion~\cite{vmvf,bpnet} 
and model ensembling~\cite{hansen1990neural,dietterich2000ensemble} techniques. On the one hand, 
multi-modal fusion takes advantage of both point clouds and images. Namely, point clouds can 
provide precise position information, while images can supply valuable color and texture information. 
On the other hand, model ensembling merges predictions from multiple models, which can be regarded 
as the voting mechanism. It is a practical technique to boost performance and is 
thus widely used in various fields. 

However, the shortcomings of the two techniques are also apparent. Although multi-modal models can make use of complementary cues from different modalities, they suffer from huge computation and memory overheads for processing two input signals simultaneously. 
For model ensembling, the training, computation and memory costs are intolerable as they grow linearly with the increasing number of model experts. This makes model ensembling infeasible for real-world applications.

\begin{figure*}[t]
	\begin{center}
		\setlength{\fboxrule}{0pt}
		\fbox{\includegraphics[width=0.50\textwidth]{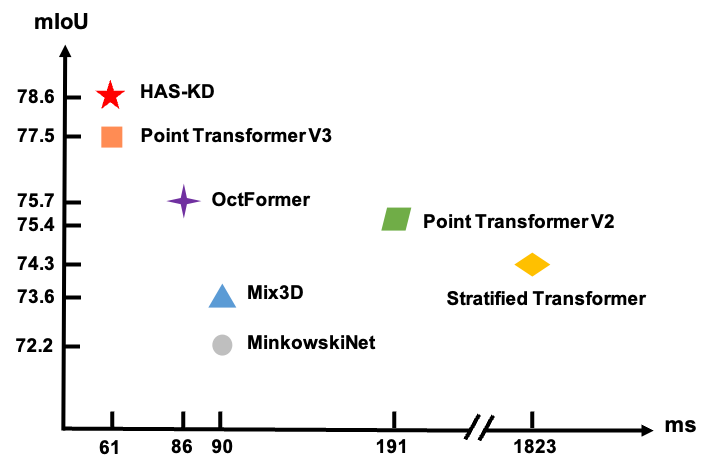}}
	\end{center}
	\vspace{-6mm}
	\caption{Comparison with state-of-the-art algorithms in terms of performance and efficiency 
 on ScanNetV2 val set. Our \algorithmname~not only achieves higher performance but also maintains 
 a lower latency compared with other competitive algorithms. The model latency is measured using one NVIDIA 4090 GPU.}
	\label{fig:efficiency}
	\vspace{-4mm}
\end{figure*}

To resolve the aforementioned dilemmas, we propose the \textbf{H}eterogeneous and \textbf{A}dept \textbf{S}napshot \textbf{K}nowledge \textbf{D}istillation framework, termed \textbf{\algorithmname}, which makes full use of knowledge distillation to transfer comprehensive information from the cumbersome multi-modal model and multiple model experts to the single-modal model. 
The single-modal model exhibits several appealing properties, including fast inference speed, 
low memory consumption and good segmentation performance. As shown in Fig.~\ref{fig:efficiency}, 
our method achieves higher performance while maintaining lower latency compared with other top-performing segmentors, such as Point Transformer V2~\cite{ptv2} and OctFormer~\cite{octformer}.

Our \algorithmname~consists of two key designs, \ie, Information-oriented Heterogeneous Distillation (IHD) and Adept Snapshot Distillation (ASD), which aim to transfer the rich knowledge from the multi-modal model and multiple experts, respectively. The IHD forces a single-modal network to produce similar features as a multi-modal network by means of knowledge distillation. It enables the model to utilize multi-modal data only during the training stage. 
Once trained, the model can perform segmentation without images and achieve higher performance than 
the original model. This is appealing because there is no extra cost at the inference stage. 
Besides, considering that a stronger teacher can teach a better student, we design an Information-Oriented Filtering (IOF) strategy to build a top-performing multi-modal teacher. Specifically, we observe that an object 
in the scene could be captured many times as the camera moves forward. Previous multi-modal methods 
\cite{bpnet,deepviewagg} randomly select several images from the image sequence, which has two shortcomings. 
First, this strategy can not capture a good view for many objects in the scene. In the selected images, 
they may appear on the boundary or be occluded heavily by other objects. Second, random selection can not 
ensure a large overlap between point clouds and images, which may constrain the image branch from supplying 
sufficient texture information to the point cloud branch. Concretely, the IOF policy tends to select the image for each object that contains the largest semantic abundance 
of the object. By projecting the points of an object to the image, we define its semantic abundance as the 
percentage of its points that appear in the view. The points of an object can be obtained by its ground truth 
instance mask, which is accessible during training. The instance mask can also be generated by contemporary 
instance segmentation models, such as \cite{sam3d}, when there are no ground truth instance masks.

To further endow a single model with the rich knowledge of multiple model experts, we propose Adept Snapshot Distillation (ASD), which leverages knowledge distillation to teach a single student model with the ensemble features 
or ensemble predictions from multiple teacher models (multi-teachers). As opposed to vanilla model ensembling that requires the separate training of each model, ASD treats the freely available model snapshots generated during the training phase as multiple experts and merely requires training the model once, which remarkably reduces the training cost for model ensembling. 
Besides, since different model experts may excel or underperform in the same pattern, utilizing the knowledge of teachers who are not proficient in the pattern can be detrimental to knowledge distillation. To this end, our ASD proposes to 
assign only an expert model for the same kind of pattern. We find that a single segmentation model can not perform the best in all classes. Therefore, we select the best-performing checkpoint for each class as the expert teacher of the corresponding class when training teacher models. By transferring knowledge from expert teachers to the student, the student is taught to perform proficiently in different classes.

Experimental results on prevalent 3D segmentation benchmarks reveal that both IHD and ASD are beneficial 
in improving the performance of the 3D segmentation model. To summarize, our contributions can be listed as follows:

\begin{itemize}
    \item We design the Information-oriented Heterogeneous Distillation (IHD), which distills the rich knowledge of a multi-modal model to a single-modal model. IHD utilizes an Information-Oriented Filtering (IOF) strategy to  
    improve the performance of the multi-modal teacher, which can benefit the learning of the student.
    \vspace{2mm}
    
    \item We devise the Adept Snapshot Distillation (ASD), which integrates the expert knowledge of 
    multiple expert teachers and transfers the knowledge to a student. These experts are freely available model snapshots produced during the training phase. Each expert provides supervision to the student in the class where it is adept at.
    \vspace{2mm}
    
    \item Our approach delivers the new state-of-the-art performance on two commonly used 3D semantic 
    segmentation benchmarks, \textit{i.e.}, ScanNetV2~\cite{scannet} and S3DIS~\cite{s3dis}. Extensive experiments on the two datasets demonstrate the effectiveness of our method.
\end{itemize}

\section{Related Work}
\label{sec:relatedwork}

\subsection{3D Semantic Segmentation}
3D semantic segmentation aims to assign a categorical label to each point in the scanned scene~\cite{taseg,3d-survey,moe3d,sega3d}. 
According to the used point cloud representations, prior works can be classified into three categories, 
including projection-based, voxel-based, and point-based methods.
Projection-based methods~\cite{lawin2017deep,boulch2017unstructured,wu2018squeezeseg,wu2019squeezesegv2} 
project 3D point clouds onto the 2D plane and employ 2D models to process the 2D projections. 
Voxel-based methods~\cite{maturana2015voxnet,song2017semantic} split irregular point clouds into regular 
voxels and process these voxels with the efficient sparse convolution~\cite{minkunet,graham20183d}.
Point-based methods~\cite{pointnet,pointnet++,ma2022rethinking, pointweb,ptv2,ptv3} extract features from 
raw or grouped point clouds. Though the above-mentioned approaches have achieved great success, 
they face limitations due to the lack of fine-grained 2D appearance and texture information.
Multi-modal fusion \cite{dai20183dmv,jaritz2019multi,huang2019texturenet,bpnet,deepviewagg,mate} 
is an emerging trend to fully exploit the potential of multiple complementary data representations.
Nevertheless, multi-modal approaches require large computational resources to handle multiple input signals 
at the same time. Besides, model ensembling \cite{dietterich2000ensemble,hansen1990neural,opitz1999popular} 
is also a commonly used technique to improve 3D semantic segmentation performance.
It combines predictions from multiple models or multiple data augmentation to make a better prediction.
The total computation overhead can be even more than that of multi-modal approaches due to multiple forwards.
In this paper, our \algorithmname~makes a good trade-off between accuracy and efficiency by transferring the 
comprehensive knowledge from the multi-modal model and ensemble model to a single-modal model. 
Eventually, our method achieves higher performance without increasing the inference burden.

\subsection{Knowledge Distillation}
Knowledge distillation (KD) stems from the seminal work of G. Hinton \etal~\cite{kd}. 
The objective of KD is to transfer the dark knowledge from the teacher model to the student network to mitigate the performance gap between these two models~\cite{fitnet,sad}. 
Recently, knowledge distillation has been widely applied to 3D scene understanding, including 3D object detection \cite{sparsekd} and 3D semantic segmentation \cite{pvkd}.
Some cross-modal knowledge distillation methods are also proposed \cite{s2m2-ssd,2dpass,unidistill}. For example, 2DPASS~\cite{2dpass} designs the multi-scale fusion-to-single knowledge distillation 
framework for cross-modal distillation. S2M2-SSD~\cite{s2m2-ssd} divides the distillation stage into 
four different objectives and boosts the performance of the single-modal 3D detector. 
Despite their impressive results, they only leverage images that synchronize with point clouds, which 
limits them from leveraging historical or future images for cross-modal distillation.
Our IHD selects a good view for each object from a long-term image sequence, which 
can supply the more informative image features for knowledge distillation.
Apart from the cross-modal distillation, multi-teacher distillation is also a useful knowledge distillation technique.
Rather than learning from a single teacher, multi-teacher knowledge distillation incorporates predictions from multiple teachers,
as the wisdom of the masses exceeds that of the wisest individual.
Previous literatures usually leverage the predictions of all teachers, such as simply assigning average or 
other fixed weights for different teachers~\cite{wu2019multi,you2017learning} and calculating the teacher weights 
based on entropy~\cite{kwon2020adaptive}, latent factor~\cite{liu2020adaptive}, uncertainty~\cite{zhang2023avatar} 
or other learnable strategies~\cite{zhang2022confidence}.
Different from them, we observe that the performance of different models on the same pattern can be different.
Leveraging the knowledge of teachers who do not excel in the pattern can be harmful.
Our ASD chooses a best-performing checkpoint for each class, which is considered the expert teacher of the class.
During training, for voxels of a class, we only distill them with the expert teacher who performs the best in the class.
In this way, the student can receive more professional teaching in different classes. 

\section{Methodology}
\label{sec:methodology}

In this paper, we take full advantage of knowledge distillation to boost the performance of the model 
without introducing extra computation and memory costs during the inference phase.
To this end, we propose two effective knowledge distillation techniques: Information-oriented Heterogeneous Distillation (IHD) and Adept Snapshot Distillation (ASD).
The detailed illustrations are provided in the following sections.

\begin{figure*}[t]
	\begin{center}
		\setlength{\fboxrule}{0pt}
        \fbox{\includegraphics[width=1.0\textwidth]{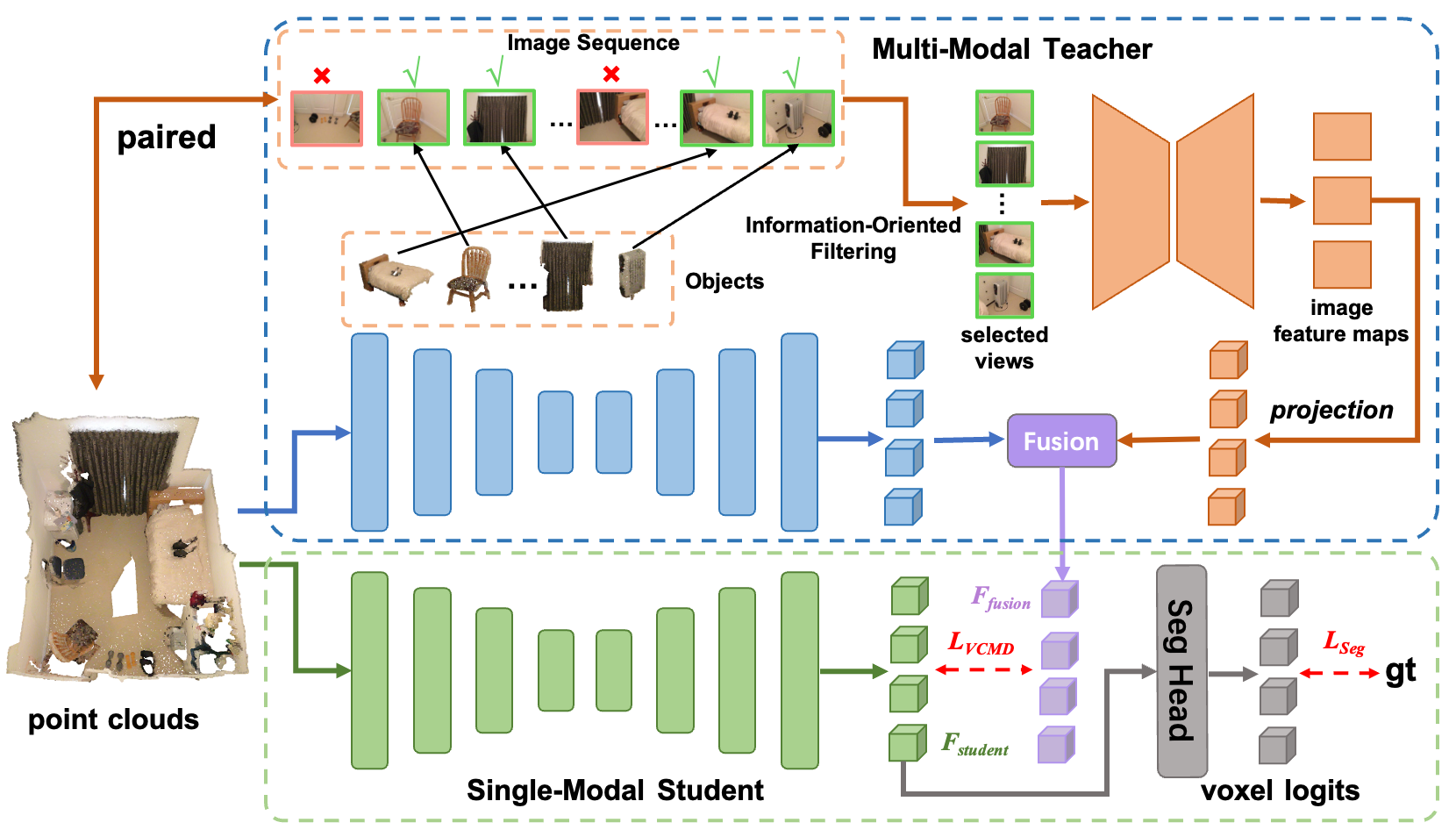}}
	\end{center}
	\vspace{-6mm}
	\caption{Illustration of our Information-oriented Heterogeneous Distillation (IHD). 
 It transfers the comprehensive knowledge of a multi-modal teacher to a single-modal student, which boosts the student without extra computation during the inference. Moreover, we present the Information-Oriented Filtering (IOF) strategy to select the most informative view for each object in the scene, which can greatly improve the multi-modal teacher and thus benefit the student. The training consists of two stages. We first pre-train a multi-modal teacher. Then, we perform knowledge distillation between the pre-trained multi-modal teacher and a single-modal student.}
	\label{fig:segkd1}
	\vspace{-4mm}
\end{figure*}

\subsection{Information-oriented Heterogeneous Distillation}
\label{sec:vcmd}
To enjoy the inference efficiency of the single-modal model while achieving good performance as the multi-modal model, 
we propose the Information-oriented Heterogeneous Distillation (IHD). IHD distills the informative knowledge from the
multi-modal model to a single-modal model. To promote cross-modal distillation, we design an Information-Oriented Filtering (IOF) strategy to boost the performance of the multi-modal teacher.

\begin{figure*}[t]
	\begin{center}
		\setlength{\fboxrule}{0pt}
        \fbox{\includegraphics[width=0.98\textwidth]{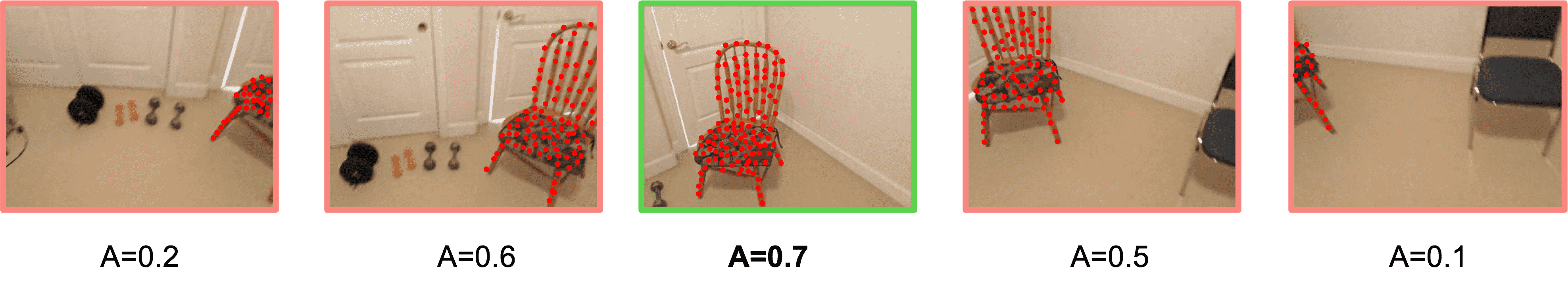}}
	\end{center}
	\vspace{-7mm}
	\caption{Illustration of Information-Oriented Filtering (IOF) strategy. We select the view that holds the highest semantic information for the chair.}
	\label{fig:motivation}
	\vspace{-6mm}
\end{figure*}

\vspace{2mm}
\noindent \textbf{Information-Oriented Filtering.}
Our Information-Oriented Filtering strategy is based on the observation that an object can appear multiple times 
in the continuous image sequence of a scene, as shown in Fig.~\ref{fig:motivation}. 
Since an image sequence usually contains hundreds of views, it is infeasible to utilize all views for multi-model fusion.
To solve this, previous multi-modal fusion methods \cite{bpnet,deepviewagg} randomly select several views from the image sequence.
However, the random selection strategy can not capture a good view of each object in the scene, 
which greatly limits the performance of multi-modal methods. 
Considering that in cross-modal distillation, the multi-modal fusion is only performed during training.
This enables us to leverage ground truth or offline priors to guide the view selection, thus improving multi-modal fusion.

Specifically, for the $i^{th}$ object in a scene, we can get its point clouds $P_i \in \mathbb{R}^{N \times 3}$ 
with its ground-truth instance labels or offline generated instance predictions from existing instance segmentation 
model, \textit{e.g.}, SAM3D \cite{sam3d}. Here, $N$ is the number of points of the object.
For the $j^{th}$ view in the scene, the dataset provides its camera matrix $T_j$ and ground truth depth map 
$D_j \in \mathbb{R}^{H \times W}$. We first project $P_i$ to the $j^{th}$ view with the camera matrix $T_j$ and 
calculate the 2D coordinates $(U_i \in \mathbb{R}^{N}, V_i \in \mathbb{R}^{N})$ and the projected depth $Z_i \in \mathbb{R}^{N}$ 
of the object on the view, which can be formulated as follows:

\vspace{-2mm}
\begin{equation}
\begin{aligned}
[U_i, V_i, Z_i] = P_i \cdot T_j.
\end{aligned}
\end{equation}

Then, we select the points of $P_i$ that are located in the view according to their 2D coordinates, 
resulting in $\widetilde{U_i}, \widetilde{V_i}, \widetilde{Z_i}$. However, not all of these points 
are visible in the view due to the occlusion caused by other objects or self. To this end, we further select the point whose projected depth $\widetilde{Z_i}$ 
is close to the ground truth depth of where they are located on the $D_j$.
By default, if the difference between the two depths is less than 2 cm, we consider them close.
Finally, we get all visible points of the $i^{th}$ object on the $j^{th}$ view, $Q_i \in \mathbb{R}^{M \times 3}$, 
where $M$ is the number of visible points. Given  $P_i \in \mathbb{R}^{N \times 3}$ and $Q_i \in \mathbb{R}^{M \times 3}$, 
we define the semantic abundance of the $i^{th}$ object on the $j^{th}$ view as the percentage of visible points of the object in the view:

\vspace{-2mm}
\begin{equation}
\begin{aligned}
A_i^j = \frac{M}{N}.
\end{aligned}
\end{equation}

Given the semantic abundance of an object on a view, we can basically judge whether the view is good or bad for the object.
A larger semantic abundance indicates that the view can capture a more complete appearance of the object.
Thus, the view can supply richer texture and color clues of the object for multi-modal fusion.
Based on the above analysis, we present the \textit{Information-Oriented Filtering} strategy.
Specifically, for each object in a scene, we just need to select the view that has the largest semantic abundance with the object for multi-modal fusion. As shown in the top left of Fig.~\ref{fig:segkd1}, we select the most informative view for each object in the scene and feed the selected views to the multi-modal model.
The experiments in our ablation study show that our view-filtering strategy can improve the performance of the multi-modal teacher largely and distill a better single-modal student.

\vspace{2mm}
\noindent \textbf{Knowledge Distillation.}
As shown in the bottom of Fig.~\ref{fig:segkd1}, we utilize the fused features $F_{\text{fusion}}$ of the multi-modal teacher to supervise the single-modal features $F_{\text{student}}$ of the student and we simply employ an MSE loss for our IHD:
\begin{equation}
\begin{aligned}
\mathcal{L}_{\text{IHD}} = \mathbb{E}[\| F_{\text{student}} - F_{\text{fusion}} \|_2].
\end{aligned}
\end{equation}

\begin{figure*}[t]
	\begin{center}
		\setlength{\fboxrule}{0pt}
		\fbox{\includegraphics[width=1.0\textwidth]{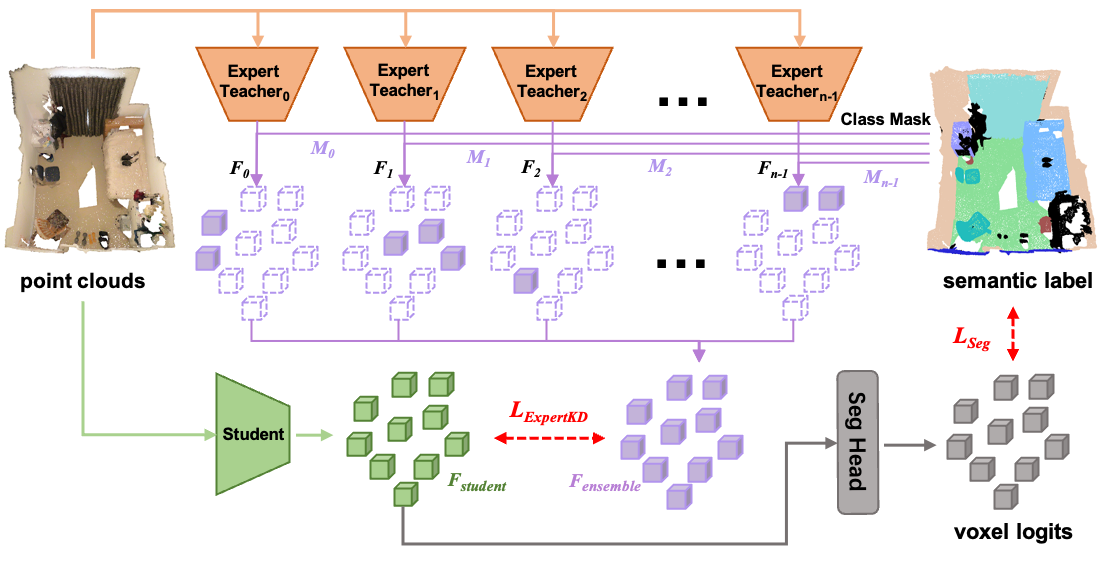}}
	\end{center}
	\vspace{-4mm}
	\caption{Illustration of our Adept Snapshot Distillation (ASD). ASD transfers the knowledge 
 of multiple expert teachers to the student. A single teacher typically can not excel in all classes and thus 
 can not provide the best supervision for the student. ASD selects the best-performing checkpoint of each class as the expert teacher of each class. Each expert teacher only contributes knowledge they are proficient in, which provides high-quality supervision in each class for the student.}
	\label{fig:segkd2}
	\vspace{-5mm}
\end{figure*}

\subsection{Adept Snapshot Distillation}
\label{sec:expertkd}
Previous multi-teacher distillation methods typically ensemble predictions of all teachers.
However, different teachers can perform differently on the same point cloud pattern.
Utilizing predictions of teachers who do not excel in the pattern can introduce noisy supervision signals to the student.
Our Adept Snapshot Distillation (ASD) first obtains multiple expert teachers who excel in different classes.
Then, for voxels that belong to a class, ASD only distills the student with the expert teacher who excels in the class.
A detailed introduction to ASD is given as follows.

\vspace{2mm}
\noindent \textbf{Expert Teachers.} 
Different from conventional model ensembling methods that require separate training for each model, 
our expert teachers can be obtained by training a model only once.
The expert teacher for the $i^{th}$ class is defined as the checkpoint that holds the highest evaluation 
results on the $i^{th}$ class. Specifically, we save the best-performing checkpoint of each class after 
each epoch during training. After the training, the best-performing checkpoint of each class is the expert 
teacher of each class. Each expert teacher performs well in the specific class, which can provide more 
professional supervision for voxels belonging to the class. Our experimental result shows that a single 
expert teacher can achieve a higher IoU than an average-based ensemble teacher in the corresponding class.

\vspace{2mm}
\noindent \textbf{Knowledge Distillation.}
With pre-trained expert teachers, we can perform our ASD, as shown in Fig.~\ref{fig:segkd2}.
Given the expert teacher $E_{i}$ for the $i^{th}$ class, we first feed the point cloud to $E_{i}$, 
resulting in the expert features $F_{i} \in \mathbb{R}^{N \times C}$, 
where N is the voxel number of the last feature map and C is the channel of voxel features.
The key idea of our ASD is that the expert teacher of the $i^{th}$ class only provides 
supervision on the voxel that belongs to the $i^{th}$ class. Therefore, we select voxel features in $F_{i}$ 
that belong to $i^{th}$ class by the class mask, $M_{i} \in \mathbb{R}^{N}$, which can be generated by 
the semantic labels of point clouds. By integrating voxel features from different expert teachers, 
we form the ensemble features $F_{\text{ensemble}}$, as shown in Equation \ref{eq:ensemble}:

\begin{equation}
\begin{aligned}
    F_{\text{ensemble}} &= F_{0}M_{0} \cup F_{1}M_{1} \cup \dots \cup F_{n-1}M_{n-1}.\\
\end{aligned}
\label{eq:ensemble}
\end{equation}

\noindent Assuming the features of the student is $F_{\text{student}} \in \mathbb{R}^{N \times C}$, 
we simply use an MSE loss between $F_{\text{student}}$ and $F_{\text{ensemble}}$ for expert knowledge distillation:
\begin{equation}
\begin{aligned}
\mathcal{L}_{\text{ASD}} = \mathbb{E}[\| F_{\text{student}} - F_{\text{ensemble}} \|_2].
\end{aligned}
\end{equation}

\subsection{Training Objective}
To improve the student model with IHD and ASD simultaneously, we replace the network of teacher models 
in ASD from a single-modal network to a multi-modal network built by IHD, and we define the unified 
knowledge distillation loss based on $\mathcal{L}_{\text{ASD}}$ as $\mathcal{L}_{\text{ASD-IHD}}$.
Then, the final training objective can be formulated as follows: 
\vspace{-2mm}
\begin{equation}
\begin{aligned}
\mathcal{L} = \mathcal{L}_{\text{Seg}} + 
\alpha \mathcal{L}_{\text{ASD-IHD}},
\vspace{-2mm}
\end{aligned}
\end{equation}
where $\alpha$ is the coefficient to control the effect of knowledge distillation losses. We set $\alpha=4$ by default.

\section{Experiments}
\label{sec:experiments}
\noindent\textbf{Datasets \& Metrics}.
We validate the effectiveness and efficiency of our \algorithmname~on ScanNetV2~\cite{scannet} and S3DIS~\cite{s3dis}.
ScanNetV2 contains 1, 513 large-scale 3D scans reconstructed from RGB-D frames.
There are 1, 201 scanned scenes for training and 312 scenes for validation.
The point clouds are sampled from vertices of reconstructed meshes, 
and each sampled point is assigned a semantic label from 20 categories, \emph{e.g.}, wall, floor, cabinet, \emph{etc}.
The S3DIS dataset consists of 271 rooms in six areas from three different buildings.
Following the experiment setting used in previous works~\cite{segcloud,pointnet++,pt}, we take area 5 as the test set and use the remaining data for training.
Different from ScanNetV2, points of S3DIS are densely sampled on the mesh surfaces and annotated with a semantic label from 13 categories.
Following the common protocol~\cite{pointnet++}, we take mean class-wise intersection over union (mIoU) 
as the evaluation metric for ScanNetV2.
We use mean class-wise intersection over union (mIoU), mean of class-wise accuracy (mAcc), 
and overall point-wise accuracy (OA) for evaluation on S3DIS Area 5.

\vspace{3mm}
\noindent\textbf{Implementation Details}.
Our experiment is based on Pointcept~\cite{pointcept} and we take Point Transformer V3 (PTV3) ~\cite{ptv3} as the baseline model.
We employ the AdamW optimizer~\cite{adamw} and the OneCycle learning rate scheduler~\cite{onecycle}. 
We train the network with a batch size of 24, a maximum learning rate of 0.006 and a weight decay of 0.05.
The training of \algorithmname~includes two phases: pre-training teacher and distilling student.
For the ScanNetV2 dataset, we first pre-train the teacher with 800 epochs and conduct knowledge distillation for 160 epochs.
For the S3DIS dataset, the epochs of pre-training and knowledge distillation are 3, 000 and 1, 200, respectively.
We use a voxel size of 2 cm for both datasets.
The initial features on ScanNetV2 include coordinates, normals and colors. The initial features on S3DIS are coordinates and colors.
The data augmentations include random dropout, random rotation, random scaling, random flip, random jitter, 
elastic distortion~\cite{minkunet}, spherical cropping~\cite{stratified} and Mix3D~\cite{mix3d}.
The probability of Mix3D is set to 0.8.
The segmentation loss consists of cross-entropy loss and Lovasz-softmax loss.
Following our baseline PTV3, we report the performance after the voting strategy.
For IHD, the maximum number of views for a scene is set to 12. 
The image branch leverages a simple UNet architecture, which produces 4 feature maps with different resolutions.
We project voxels of the backbone model to image feature maps with resolutions of 1/4 and 1/1 to gather multi-scale image features.
The voxel features and gathered image features are concatenated to produce fusion features.
For ASD without IHD, the teacher model shares the same architecture as the student model.
By pre-training the teacher model once, we can get all expert teachers as mentioned in Section \ref{sec:expertkd}.
For ASD with IHD, the teacher model is the multi-modal model with the proposed IOF strategy.

\begin{table}[t]
\centering
\renewcommand\tabcolsep{4pt}
\caption{Semantic segmentation results on ScanNetV2 val set. \algorithmname~surpasses all methods that do not use multi-modal fusion (M.M.F) and multi-dataset fusion (M.D.F). Besides, \algorithmname~achieves comparable results with Point Transformer V3$^{\dag}$. Here, $\dag$ represents that the model utilizes multiple datasets training, which pre-trains the model on the Structure3D (about 17 times larger than ScanNetV2) before training on the ScanNetV2.}
    \resizebox{.99\linewidth}{!}{
        \begin{tabular}{l|c|c|c}
          \hline
          Method                                      & M.M.F. & M.D.F. & mIoU      \\ 
          \hline
          PointNet++~\cite{pointnet++}                & $\times$ & $\times$ & 53.5     \\ 
          PointASNL~\cite{pointasnl}                  & $\times$ & $\times$ & 63.5     \\ 
          KPConv~\cite{kpconv}                        & $\times$ & $\times$ & 69.2     \\ 
          SparseConvNet~\cite{sparseconvnet}          & $\times$ & $\times$ & 69.3     \\ 
          Point Transformer~\cite{pt}                 & $\times$ & $\times$ & 70.6     \\ 
          Fast Point Transformer~\cite{fast-pt}       & $\times$ & $\times$ & 72.1     \\ 
          MinkowskiNet~\cite{minkunet}                & $\times$ & $\times$ & 72.2     \\ 
          LargeKernel3D~\cite{largekernel3d}          & $\times$ & $\times$ & 73.2     \\ 
          Mix3D~\cite{mix3d}                          & $\times$ & $\times$ & 73.6     \\ 
          Stratified Transformer~\cite{stratified}    & $\times$ & $\times$ & 74.3     \\ 
          O-CNN~\cite{ocnn}                          & $\times$ & $\times$ & 74.5     \\ 
          Point Transformer V2~\cite{ptv2}            & $\times$ & $\times$ & 75.4     \\ 
          OctFormer~\cite{octformer}                  & $\times$ & $\times$ & 75.7     \\ 
          Swin3D~\cite{swin3d}                        & $\times$ & $\times$ & 76.4     \\ 
          \underline{Point Transformer V3}~\cite{ptv3}            & $\times$ & $\times$ & \underline{77.5}     \\ 
          \textbf{\algorithmname~(Ours)}                                & $\times$ & $\times$ & \textbf{78.6}     \\ 
          \hline
          BPNet~\cite{bpnet}                          & \checkmark & $\times$ & 69.7     \\ 
          DeepViewAgg\cite{deepviewagg}               & \checkmark & $\times$ & 71.0     \\ 
          VMVF~\cite{vmvf}                            & \checkmark & $\times$ & 76.4     \\ 
          Swin3D$^\dag$~\cite{swin3d}                 & $\times$ & \checkmark & 77.5     \\ 
          Point Transformer V3$^\dag$~\cite{ptv3}     & $\times$ & \checkmark & \textbf{78.6} \\ 
          \hline
        \end{tabular}
    }
\vspace{-3mm}
\label{tab:scannetv2}
\end{table}

\subsection{Comparison with State-of-the-arts}
We compare our \algorithmname~with a series of previous state-of-the-art methods on ScanNetV2 and S3DIS, as summarized in Table~\ref{tab:scannetv2} and Table~\ref{tab:s3dis}. 
For ScanNetV2, our \algorithmname~achieves the best performance among all methods that do not adopt multi-dataset training.
Specifically, \algorithmname~outperforms Point Transformer V3 (PTV3)~\cite{ptv3} by 1.1 mIoU and OctFormer~\cite{octformer} by 2.9 mIoU.
Besides, \algorithmname~surpasses MinkowskiNet~\cite{minkunet} by 6.4 mIoU, LargeKernel3D~\cite{largekernel3d} by 5.4 mIoU, StratifiedFormer~\cite{stratified} by 4.3 mIoU and Swin3D~\cite{swin3d} by 2.2 mIoU.
The multi-dataset training for 3D semantic segmentation is utilized in ~\cite{swin3d,ppt,ptv3}, which first pre-trains the model on the Structured3D (about 17 times larger than ScanNet) and uses the pre-trained weight for fine-tuning on ScanNet. Our \algorithmname~attains the comparable result with PTV3 $^\dag$~\cite{ptv3} trained with multiple datasets, which verifies the data efficiency of our method.
For S3DIS, our \algorithmname~also achieves a good performance, as shown in Table \ref{tab:s3dis}. The result of PTV3 on S3DIS is not stable (see the issue of PTV3 codes), and we report the highest result after training PTV3 multiple times for a fair comparison between our \algorithmname~and PTV3. The table shows that our \algorithmname~improves our high-performance baseline, PTV3, by 1.0 mIoU. The encouraging results on ScanNet and S3DIS demonstrate the effectiveness of our approach.

\begin{table}[t]
\centering
\renewcommand\tabcolsep{4pt}
\caption{Semantic segmentation results on S3DIS Area 5. $\ddag$ represents the result re-produced by us. Our method achieves the state-of-the-art result on S3DIS.}
\vspace{-1mm}
    \resizebox{.91\linewidth}{!}{
        \begin{tabular}{l | c | c | c}
            \hline
            Method                       & mIoU & OA            & mAcc \\
            \hline
            PointNet~\cite{pointnet}     & 41.1 & -             & 49.0 \\
            SegCloud~\cite{segcloud}     & 48.9 & -             & 57.4 \\
            TanConv~\cite{tanconv}       & 52.6 & -             & 62.2 \\
            PointCNN~\cite{pointcnn}     & 57.3 & 85.9          & 63.9 \\
            PointWeb~\cite{pointweb}     & 60.3 & 87.0          & 66.6 \\
            HPEIN~\cite{hpein}           & 61.9 & 87.2          & 68.3 \\
            GACNet~\cite{gacnet}         & 62.9 & 87.8          & -    \\
            SegGCN~\cite{seggcn}         & 63.6 & 88.2          & 70.4 \\
            MinkowskiNet~\cite{minkunet} & 65.4 & -             & 71.7 \\
            PAConv~\cite{paconv}         & 66.6 & -             & -    \\
            KPConv~\cite{kpconv}         & 67.1 & -             & 72.8 \\
            Point Transformer~\cite{pt}                & 70.4 & 90.8 & 76.5 \\
            Point Transformer V2~\cite{ptv2}           & 71.6 & 91.1 & \textbf{77.9} \\
            \underline{Point Transformer V3$^{\ddag}$}~\cite{ptv3} & \underline{71.1} & \underline{91.3} & \underline{76.2} \\
            \hline
            \textbf{\algorithmname~(Ours)}                               & \textbf{72.1} & \textbf{91.4} & 77.8 \\
            \hline
        \end{tabular}
    }
\label{tab:s3dis}
\end{table}

\subsection{Ablation Studies} \label{subsec:ablation}
We conduct ablation studies to examine the effectiveness of each component of our method. 
The performance is evaluated on the ScanNetV2 validation set, and we take mIoU as the evaluation metric.

\begin{table}[t]
    \caption{Ablation study of each component of our \algorithmname~on ScanNetV2 val set. The baseline is PTV3\cite{ptv3}. The results show that both IHD and ASD can improve the baseline. Leveraging them together can further improve the performance.}
    \centering
    \renewcommand\tabcolsep{8pt}
    \resizebox{.82\linewidth}{!}{
    \begin{tabular}{c|ccc|c}
        \toprule
        ID  & Baseline  & IHD   & ASD   & mIoU \\ \hline
        (1) & \checkmark &            &            & 77.5 \\
        (2) & \checkmark & \checkmark  &            & 77.9 \\
        (3) & \checkmark &            & \checkmark  & 78.0 \\
        (4) & \checkmark & \checkmark  & \checkmark  & \textbf{78.6} \\
        \bottomrule 
    \end{tabular}
    }
    \label{exp:components}
    \vspace{-2mm}
\end{table}

\vspace{2mm}
\noindent \textbf{Ablation on IHD and ASD}.
The effect of the proposed Information-oriented Heterogeneous Distillation (IHD) and Adept Snapshot Distillation (ASD) is summarized in Table~\ref{exp:components}. The baseline is chosen as PTV3~\cite{ptv3}, which is the current state-of-the-art method. When applying IHD, the performance is improved from 77.5 to 77.9. Besides, ASD brings a 0.6 mIoU gain, boosting the performance from 77.4 to 78.0. 
It is worth noting that PTV3 is a top-performing segmentation model, and bringing gains to such a competitive baseline is very challenging. Therefore, the improvements achieved by IHD and ASD are considerable. 
Furthermore, by leveraging both IHD and ASD, we achieved a 1.1 mIoU gain, elevating the baseline performance from 77.4 to 78.6. The final performance exceeds all previous methods 
and achieves comparable performance to the method trained with multiple datasets, 
which confirms the superiority of our method.

\begin{table*}[ht]
    \begin{center}
    \noindent\begin{minipage}[t]{.48\linewidth}
    \centering
        \caption{Ablation study on the view-filtering strategy for cross-modal knowledge distillation. 
        We report the performance of the multi-modal teacher and the performance of the single-modal student after cross-modal distillation.}
        \label{exp:view}
        \centering
            \begin{tabular}{c|cc}
                \toprule
                View-Filtering Strategy & Teacher        & Student \\ \hline
                Random Filtering            & 78.1           & 77.5 \\
                Information-Oriented Filtering            & \textbf{78.8}  & \textbf{77.9} \\
                \bottomrule 
            \end{tabular}
    \end{minipage}\hfill
    \centering
    \noindent\begin{minipage}[t]{.48\linewidth}
    \centering
        \caption{Ablation study on the ensemble strategy for multi-teacher knowledge distillation. 
        We report the performance of the ensemble teacher and the performance of the single student after multi-teacher knowledge distillation.}
        \label{exp:ensemble}
        \centering
            \begin{tabular}{c|cc}
                \toprule
                Ensemble Strategy & Teacher        & Student\\ \hline
                Average Ensemble          & 77.9           & 77.6 \\
                Expert Ensemble           & \textbf{79.0}  & \textbf{78.0} \\
                \bottomrule 
            \end{tabular}
    \end{minipage}
    \end{center}
    \vspace{-5mm}
\end{table*}

\vspace{2mm}
\noindent \textbf{Ablation on Information-Oriented Filtering}.
We conduct an ablation study to explore the effect of different view-filtering strategies 
for cross-modal knowledge distillation, and the results are recorded in Table~\ref{exp:view}.
Due to the memory constraint, it is impractical to feed all views of a scene to the multi-modal models, 
necessitating the introduction of a view-filtering strategy.
Previous multi-modal methods randomly select a number of views for multi-modal fusion.
However, as mentioned in Section \ref{sec:vcmd}, this Random Filtering (RF) strategy can not ensure the image quality for objects in the scene.
The proposed Information-Oriented Filtering (IOF) chooses the view that holds the high semantic abundance for each object to improve the multi-modal fusion.
The experiment results show that our IOF can lead to a higher-performing multi-modal teacher 
than RF (78.8 v.s. 78.1). Typically, a better teacher can teach a better student. The performance of students distilled by different multi-modal teachers in the table confirms this hypothesis.
Specifically, our IOF outperforms RF by 0.4 mIoU on the student, which showcases the effectiveness of the IOF.

\vspace{2mm}
\noindent \textbf{Ablation on Multi-Teacher Ensemble Strategy}.
We investigate the effect of different ensemble strategies for multi-teacher knowledge distillation.
The comparisons are presented in Table \ref{exp:ensemble}.
``Average Ensemble'' represents the commonly used ensemble strategy that utilizes the average of predictions from multiple teachers to distill the student. ``Expert Ensemble'' represents our method that selects the best-performing checkpoint in each class as the expert teacher of the class. Each expert teacher only supplies its knowledge on the class it performs well. The experiment results show that our method can produce a higher-performance ensemble teacher, which yields
79.0 mIoU and surpasses the average-based ensemble teacher by 1.1 mIoU. As a result, the expert ensemble strategy also teaches a better student (78.0 v.s. 77.6). The improvement can be attributed to the expert knowledge provided by expert teachers. The average-based ensemble teacher can introduce noise knowledge from underperformed teachers and our method can alleviate this.

\begin{table}[t]
    \caption{Comparison on model parameters, inference latency and memory with other methods. The latency and memory are measured on an NVIDIA 4090 GPU.}
    \centering
    \renewcommand\tabcolsep{6pt}
    \resizebox{1.0\linewidth}{!}{
        \begin{tabular}{l|rrr}
        \toprule
        Method                     & \#Params  & Latency & Memory  \\ \hline
        MinkUNet\cite{minkunet}    & 37.9M       & 90ms    & 4.7G    \\
        OctFormer\cite{octformer}  & 44.0M       & 86ms    & 12.5G   \\
        Swin3D\cite{swin3d}        & 71.1M       & 456ms   & 8.8G    \\
        Point Transformer V2 \cite{ptv2} & 12.8M & 191ms   & 18.2G   \\
        Point Transformer V3 \cite{ptv3} & 46.2M & 61ms    & 5.2G    \\
        \algorithmname~(Ours)               & 46.2M       & 61ms    & 5.2G    \\
        \bottomrule 
        \end{tabular}
    }
    \label{exp:efficiency}
    \vspace{-2mm}
\end{table}

\vspace{2mm}
\noindent \textbf{Efficiency Analysis}.
In Table~\ref{exp:efficiency}, we provide a comparison on model parameters, inference latency and memory consumption with previous state-of-the-arts, all of which are evaluated on an NVIDIA 4090 GPU.
The table shows that \algorithmname~maintains a fast inference speed, a low memory overhead and a moderate model capacity, which can be attributed to the design of our IHD and ASD, and the light-weight baseline.
IHD and ASD transfer the knowledge from the strong multi-modal teacher and expert teachers to a student, which brings a free-lunch performance boost.
Once trained, the student can do the inference with the single-modal input and single network. 
The network of the student is not modified throughout the process. 
In our method, we take PTV3 \cite{ptv3} as our student, which enables our \algorithmname~to inherit its efficiency.
Specifically, \algorithmname~achieves the lowest latency as PTV3 among the methods in the table.
Compared to PTV3, our \algorithmname~brings a 1.1\% performance gain without resorting to multi-modal fusion or model ensembling.
This improvement shows the success of our IHD and ASD.
In addition, our \algorithmname~achieves comparable results with the PTV3$^\dag$ that is trained with a large-scale extra dataset. This also demonstrates the data efficiency of our method.



\section{Conclusion}\label{conclusion}
In this paper, we fully exploit knowledge distillation to boost the performance of the 3D semantic segmentation model without introducing extra computation, memory and parameter costs at the inference stage.
The proposed knowledge distillation framework \algorithmname~consists of two key designs: Information-oriented Heterogeneous Distillation (IHD) and Adept Snapshot Distillation (ASD).
IHD and ASD transfer informative knowledge from the multi-modal teacher and expert teachers to a single-modal teacher, which achieves impressive results. Extensive experiments on ScanNet and S3DIS datasets validate the effectiveness of our method. In particular, our \algorithmname~delivers a state-of-the-art result while maintaining high inference efficiency.

\clearpage
{\small
\bibliographystyle{ieee_fullname}
\bibliography{egbib}
}

\end{document}